\documentclass[sigconf]{acmart}

\renewcommand\footnotetextcopyrightpermission[1]{}
\usepackage{fancyhdr}
\usepackage{graphicx}
\usepackage{booktabs}
\usepackage{amsmath,amsfonts}
\usepackage{microtype}
\usepackage{xcolor}
\usepackage{enumitem}
\usepackage{caption}
\usepackage{fancyhdr} 

\usepackage{adjustbox}
\usepackage{array}
\usepackage{multirow}
\usepackage{placeins}
\usepackage{float}
\usepackage{verbatimbox}

\graphicspath{{figs/}}
\DeclareGraphicsExtensions{.pdf,.png,.jpg}

\setcitestyle{numbers,sort&compress}

\title{CPEMH: An Agentic Framework for Prompt-Driven Behavior Evaluation and Assurance in Foundation-Model Systems for Mental Health Screening}

\author{Giuliano Lorenzoni}
\affiliation{
  \institution{University of Waterloo}
  \city{Waterloo}
  \state{ON}
  \country{Canada}
}
\email{glorenzo@uwaterloo.ca}

\author{Ivens Portugal}
\affiliation{
  \institution{University of Waterloo}
  \city{Waterloo}
  \state{ON}
  \country{Canada}
}
\email{iportugal@uwaterloo.ca}

\author{Paulo Alencar}
\affiliation{
  \institution{University of Waterloo}
  \city{Waterloo}
  \state{ON}
  \country{Canada}
}
\email{palencar@uwaterloo.ca}

\author{Donald Cowan}
\affiliation{
  \institution{University of Waterloo}
  \city{Waterloo}
  \state{ON}
  \country{Canada}
}
\email{dcowan@uwaterloo.ca}

\begin{document}

\begin{abstract}
This paper presents \textbf{CPEMH}, an agentic framework designed to evaluate prompt-driven behavior in foundation-model systems operating on transcript-based datasets for \textit{mental-health screening}. CPEMH serves as an \textit{engineering methodology for behavioral assurance} in large-scale language systems, introducing an orchestrated architecture that autonomously performs the \textbf{design, evaluation, and selection of prompt strategies}, enabling systematic control of behavioral variability across contexts. Its modular agentic design, combining orchestrator, inference, and evaluation agents, ensures traceability, reproducibility, and robustness throughout the prompting lifecycle. A case study on automated depression screening from interview transcripts demonstrates the framework’s capacity to stabilize and audit foundation-model behavior in conversational and clinically sensitive domains. \textbf{Lessons learned} emphasize the role of modular orchestration in behavioral assurance, the prioritization of stability over architectural complexity, and the integration of \textit{F1, bias, and robustness} as core acceptance criteria.
\end{abstract}



\maketitle
\thispagestyle{plain}
\pagestyle{plain}
\fancyhf{}

\section{Introduction}
Foundation Models (FMs) and Large Language Models (LLMs) have become integral to decision-making processes across domains such as healthcare, education, and finance. Yet, their deployment in sensitive contexts raises a fundamental challenge: behavioral assurance, the ability to guarantee stable, explainable, and reproducible behavior under prompt and data variability. In transcript-based health scenarios, where minor wording or contextual shifts may alter diagnostic outcomes, this challenge becomes critical for safety, ethics, and reliability.

Conceived as the \textit{Contextual Prompt Enabler for Mental Health (CPEMH)}, the framework serves as an agentic architecture for prompt-driven behavioral evaluation and assurance in foundation-model systems for mental health screening. CPEMH formalizes the design–evaluation–selection loop as an engineering process, transforming prompt engineering from ad hoc experimentation into a controlled and auditable pipeline. 

Recent advances in agentic AI have shown that multi-agent orchestration can extend LLM reasoning through delegation, coordination, and feedback. However, most existing systems emphasize \textit{task automation} or \textit{performance improvement} rather than behavioral reliability. In mental-health screening, where predictions influence clinical decision support, inconsistent or opaque responses undermine user trust and regulatory compliance.

To address these issues, CPEMH operationalizes behavioral assurance by (i) orchestrating modular agents for dataset preparation, inference, and evaluation, (ii) computing behavioral metrics such as bias and robustness alongside accuracy, and (iii) enforcing reproducibility through consistent orchestration across runs.

The framework’s contributions are fourfold:
\begin{itemize}[leftmargin=0.8em]
    \item A multi-agent architecture that operationalizes behavioral assurance for prompt-driven LLM systems.
    \item A metric-driven evaluation loop quantifying bias, robustness, and consistency beyond accuracy.
    \item An empirical case study on depression detection from clinical interviews, validating stability and interpretability.
    \item Engineering lessons highlighting modular orchestration, reproducibility, and metric-based governance in reliable LLM deployment.
\end{itemize}

In doing so, CPEMH bridges research in agentic systems and prompt engineering with behavioral governance, providing a blueprint for reproducible, auditable, and stable AI behavior in transcript-based mental-health screening. Results are reported to support prompt-driven behavior evaluation and assurance, with a focus on stability, bias, and governance properties, as prerequisites for scalable and production-ready foundation model systems, rather than on state-of-the-art predictive performance.

\section{Related Work}

\subsection{Agentic Systems in Healthcare}

Agentic systems are software systems composed of interacting components, LLM agents, that collaborate to achieve individual or collective goals~\cite{hughes2025agents}. 
An LLM agent uses a large language model as its reasoning core to perform tasks by interpreting instructions, planning actions, managing memory, and interacting with external tools~\cite{zhao2024expel}. 
While many agentic systems have been applied to software maintenance and code generation~\cite{dong2024selfcollab,toprani2025llmworkflow,jin2024teach}, their application is expanding to healthcare domains~\cite{qiu2024llmhealth}. 

\textit{AgentMD}~\cite{jin2024agentmd} performs clinical risk prediction through automated calculator selection, achieving over 80\% accuracy on benchmark metrics. 
\textit{MedAide}~\cite{yang2025medaide} applies cooperative multi-agent reasoning for medical intent detection, while \textit{PST-Agent}~\cite{wang2025pst} adapts Problem-Solving Therapy for mental-health support through retrieval-augmented generation (RAG) and in-context learning (ICL). 
These advances demonstrate how agentic architectures enable distributed autonomy in clinical AI but still lack mechanisms for behavioral assurance, reproducibility, and post-deployment stability—central objectives of CPEMH.

\subsection{Prompt Engineering and Behavioral Evaluation}

Prompt engineering~\cite{sasaki2024patterns} focuses on designing text instructions to guide LLM behavior toward more accurate or controllable outputs. 
Prominent approaches include zero-shot, few-shot, and Chain-of-Thought prompting~\cite{brown2020language,wei2022chain}. 
Recent research has explored structured and role-based prompts to enhance interpretability~\cite{white2024patterns,zhou2023humanprompt}. Evaluation efforts combine automatic and hybrid metrics such as BLEU, ROUGE, METEOR, and BERTScore, as well as agreement measures like Cohen’s $\kappa$~\cite{son2025optimizing,choi2025efficient}. 
Consistency and robustness metrics have also been proposed to evaluate stability across paraphrased prompts or contextual variations~\cite{zhao2021calibrate,perez2021truefewshot}. 
CPEMH extends these perspectives by embedding prompt evaluation into a multi-agent process that operationalizes behavioral assurance and transparent assessment.





.

\section{Methodology and Framework}
\subsection{Agentic Design Principles}
CPEMH follows three agentic principles:
(1) \textit{Autonomous Orchestration} — coordinated control of design–evaluation–selection loops;
(2) \textit{Traceable Evaluation} — metrics aligned to behavioral properties (bias, robustness, consistency);
(3) \textit{Behavioral Assurance} — quantifying and minimizing context-driven variability.

\subsection{Architectural Overview}
\begin{figure}[!t]
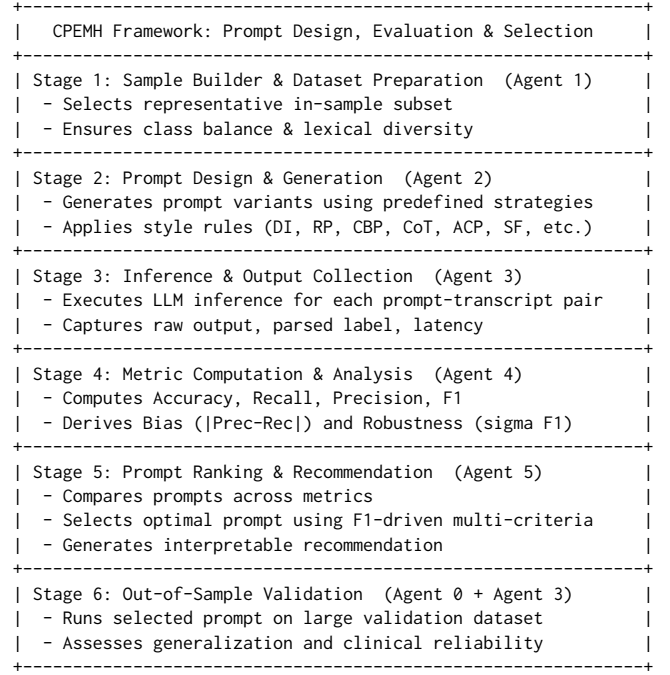

  \centering
  \resizebox{\columnwidth}{!}{\theverbbox}
  \caption{Workflow of the CPEMH framework stages and associated agents.}
  \label{fig:cpehm_workflow}
\end{figure}

Figure~\ref{fig:cpehm_workflow} illustrates the CPEMH workflow. The \textbf{Orchestrator Agent} manages prompt generation and coordination, the \textbf{Inference Agent} executes LLM queries under controlled hyper-parameters, and the \textbf{Evaluation Agent} computes performance and behavioral metrics.

\begin{verbbox}
+--------------------------------------------------------------+
|   CPEMH Framework: Prompt Design, Evaluation & Selection     |
+--------------------------------------------------------------+
| Stage 1: Sample Builder & Dataset Preparation  (Agent 1)     |
|  - Selects representative in-sample subset                   |
|  - Ensures class balance & lexical diversity                 |
+--------------------------------------------------------------+
| Stage 2: Prompt Design & Generation  (Agent 2)               |
|  - Generates prompt variants using predefined strategies     |
|  - Applies style rules (DI, RP, CBP, CoT, ACP, SF, etc.)     |
+--------------------------------------------------------------+
| Stage 3: Inference & Output Collection  (Agent 3)            |
|  - Executes LLM inference for each prompt-transcript pair    |
|  - Captures raw output, parsed label, latency                |
+--------------------------------------------------------------+
| Stage 4: Metric Computation & Analysis  (Agent 4)            |
|  - Computes Accuracy, Recall, Precision, F1                  |
|  - Derives Bias (|Prec-Rec|) and Robustness (sigma F1)       |
+--------------------------------------------------------------+
| Stage 5: Prompt Ranking & Recommendation  (Agent 5)          |
|  - Compares prompts across metrics                           |
|  - Selects optimal prompt using F1-driven multi-criteria     |
|  - Generates interpretable recommendation                    |
+--------------------------------------------------------------+
| Stage 6: Out-of-Sample Validation  (Agent 0 + Agent 3)       |
|  - Runs selected prompt on large validation dataset          |
|  - Assesses generalization and clinical reliability          |
+--------------------------------------------------------------+
\end{verbbox}

\subsection{Specialized Agents and Roles}
\label{subsec:agents}

The Contextual Prompt Enabler for Mental Health (CPEMH) implements a modular, agentic architecture where each agent is responsible for a specific phase of the workflow. This structure supports interpretability, reproducibility, and behavioral assurance—fundamental to applying foundation models in sensitive domains such as mental health screening.  

Each agent operates autonomously but communicates through structured message passing under the supervision of a global orchestrator (Agent~0). This design enables parallelism and controlled experimentation, ensuring that all evaluations of prompt strategies remain traceable and comparable across runs.  

\begin{table}[t]
\centering
\caption{\textbf{Specialized Agents in the CPEMH Framework.}}
\label{tab:agents}
\resizebox{\columnwidth}{!}{
\begin{tabular}{p{0.13\columnwidth} p{0.27\columnwidth} p{0.53\columnwidth}}
\toprule
\textbf{Agent ID} & \textbf{Module / Function} & \textbf{Primary Responsibilities} \\ 
\midrule
\textbf{Agent 0} & Orchestrator & Coordinates workflow, manages stage dependencies, triggers agent execution, and ensures configuration consistency across experiments.  \\

\textbf{Agent 1} & Dataset Manager / Sample Builder & Selects representative in-sample subset ensuring class balance and linguistic diversity; prepares datasets for downstream inference. \\

\textbf{Agent 2} & Prompt Design and Generation Agent & Generates structured prompts from predefined strategies (Direct Instruction, Role-Based, CoT, CBP, etc.); maintains task framing uniformity and enables scalable exploration of prompt space. \\

\textbf{Agent 3} & Inference Agent & Executes generated prompts using the target LLM; logs predictions, outputs, and metadata for evaluation. \\

\textbf{Agent 4} & Evaluation Agent & Computes metrics (Accuracy, Recall, Precision, F1), bias ($|Precision - Recall|$), and robustness (F1 standard deviation); saves results for comparison. \\

\textbf{Agent 5} & Selection / Recommendation Agent & Aggregates metrics, ranks strategies, identifies optimal trade-offs between performance, bias, and stability, and produces interpretable recommendations. \\
\bottomrule
\end{tabular}
}
\vspace{-0.25cm}
\end{table}

This agent distribution mirrors the workflow shown in Fig.~\ref{fig:cpehm_workflow}. Although Agent~0 (Orchestrator) is not represented as an explicit stage in the figure, it supervises and synchronizes all operations, ensuring that specialized agents operate under a unified experimental configuration. Together, this orchestration forms a reproducible and auditable experimental pipeline—essential for behavioral assurance in transcript-based health applications where reliability, transparency, and regulatory compliance are paramount.

\subsection{Evaluation Metrics}
\label{subsec:metrics}

The \textbf{CPEMH} framework evaluates both predictive and behavioral dimensions of large-language-model performance. 
Its quantitative design follows the definitions established in the original formulation of the framework~\cite{powers2020evaluation,seneviratne2022bias}. 
The objective is not only to measure accuracy but to verify the behavioral assurance of each prompting strategy—how consistently and equitably the model behaves across contextual and data variations.

\subsubsection{Predictive Performance}
Predictive performance is assessed through the classical metrics of \emph{Precision}, \emph{Recall}, and \emph{F1-score}, computed as:
\begin{equation}
\text{Precision}=\frac{TP}{TP+FP},\quad
\text{Recall}=\frac{TP}{TP+FN},
\end{equation}
\begin{equation}
F1 = \frac{2\times\text{Precision}\times\text{Recall}}{\text{Precision}+\text{Recall}},
\end{equation}
where \(TP\), \(FP\), and \(FN\) denote true positives, false positives, and false negatives, respectively.
Together, these metrics quantify the model’s ability to correctly identify depressive subjects (recall) while maintaining reliability in its predictions (precision).

\subsubsection{Bias}
Classification bias represents the imbalance between precision and recall, capturing the degree to which a prompting strategy over- or under-detects positive cases. 
For a prompt \(p_i\),
\begin{equation}
\text{Bias}(p_i)=\big|\text{Precision}(p_i)-\text{Recall}(p_i)\big|.
\end{equation}
Lower bias indicates a balanced predictive behavior—critical in mental-health screening, where false negatives carry higher ethical cost.

\subsubsection{Robustness}
Robustness measures the stability of a prompt’s performance across repeated runs or dataset partitions. 
It is defined as the standard deviation of the F1-score over \(n\) evaluations:
\begin{equation}
\text{Robustness}(p_i)=\sigma(F1_{p_i})=
\sqrt{\frac{1}{n-1}\sum_{k=1}^{n}(F1_{p_i}^{(k)}-\overline{F1_{p_i}})^2}.
\end{equation}
Smaller $\sigma(F1)$ values imply higher stability and reduced sensitivity to contextual perturbations.

\subsubsection{Consistency and Behavioral Assurance}
Beyond individual metrics, CPEMH introduces a behavioral layer evaluating inter-prompt \textit{consistency}—the degree of agreement among different prompt variants or across in-sample/out-of-sample subsets. 
High consistency indicates predictable behavior under controlled perturbations, evidencing behavioral assurance across the prompting lifecycle.

\subsubsection{Interpretation and Decision Criterion}
These three dimensions—predictive performance, bias, and robustness—form the quantitative foundation for the decision process executed by the \emph{Prompt Selection / Recommendation Agent} (Agent 5). 
Rather than applying an explicit weighted optimization, the agent adopts a metric-driven heuristic that prioritizes the highest F1-score while enforcing acceptable bias and variance thresholds. 
This approach ensures that the final prompt recommendations achieve high predictive quality without compromising fairness or stability, thereby operationalizing the principles of behavioral assurance and reproducible evaluation embedded in the CPEMH framework.

\section{Case Study: Mental-Health Screening}

The case study employs the Distress Analysis Interview Corpus – Wizard-of-Oz (DAIC-WOZ) dataset~\cite{gratch2014daic}, which comprises 189 transcribed clinical interviews labeled for depression presence. The dataset is a clinically validated resource widely used for mental health research involving depression, anxiety, and post-traumatic stress disorder.  
Additionally, the DAIC-WOZ database is part of a larger corpus and is available to the research community upon request through the official distribution portal 
\url{https://dcapswoz.ict.usc.edu}.


The data are partitioned into two subsets:
\textit{In-Sample (IS)} — a small, balanced subset designed to maximize class and lexical representativeness; and
\textit{Out-of-Sample (OOS)} — approximately four times larger, used to induce contextual variation for behavioral evaluation.

Seven prompt-strategy families are considered:
Direct Instruction (DI), Role-Based (RP), Chain-of-Thought (CoT),
Self-Consistency (SC), Counterfactual Behavioral Prompting (CBP),
Adaptive CoT (ACP), and Structured Reasoning (SR).
Each family includes two to four prompt variants, totaling 28 configurations.

All analyses reuse the metrics, tables, and figures from the original CPEMH implementation (no recomputation).
The evaluation protocol and decision heuristic follow Section~\ref{subsec:metrics} and Section~\ref{subsec:agents};
quantitative findings are reported in Section~\ref{sec:results}.



\section{Results and Lessons Learned}
\label{sec:results}


\subsection{In-Sample Design–Evaluation–Selection}
CPEMH evaluated 28 prompt variants across seven families (DI, RP, CBP, SCP, CoT, ACP, SF). 
The Evaluation Agent computed Accuracy, Precision, Recall, F1, Bias ($|P{-}R|$), and Robustness ($\sigma_{F1}$). 
Macro-F1 reached 0.57 overall, with the top-5 prompts in Table~\ref{tab:top5_prompts} clustering around 0.66–0.69.
Direct Instruction (DI) and Role-Based Prompting (RP) showed the best balance between accuracy and variance; DI-2 was selected for validation.

\begin{table}[t]
\centering
\caption{Top-5 prompts (in-sample).}
\label{tab:top5_prompts}
\resizebox{\columnwidth}{!}{
\begin{tabular}{lcccc}
\toprule
\textbf{Prompt ID} & \textbf{Approach} & \textbf{F1} & \textbf{Accuracy} & \textbf{Precision / Recall} \\
\midrule
DI-2 & Direct Instruction (DI) & \textbf{0.687} & 0.652 & 0.54 / 0.91 \\
RP-3 & Role-Based Prompting (RP) & 0.666 & 0.643 & 0.56 / 0.84 \\
CBP-2 & Constraint-Based Prompting (CBP) & 0.664 & 0.639 & 0.51 / 0.93 \\
DI-1 & Direct Instruction (DI) & 0.662 & 0.640 & 0.52 / 0.91 \\
DI-3 & Direct Instruction (DI) & 0.660 & 0.637 & 0.50 / 0.92 \\
\bottomrule
\end{tabular}}
\vspace{-0.3cm}
\end{table}

\subsection{Out-of-Sample Validation and Comparative View}
\begin{figure}[t]
\centering
\includegraphics[width=\columnwidth]{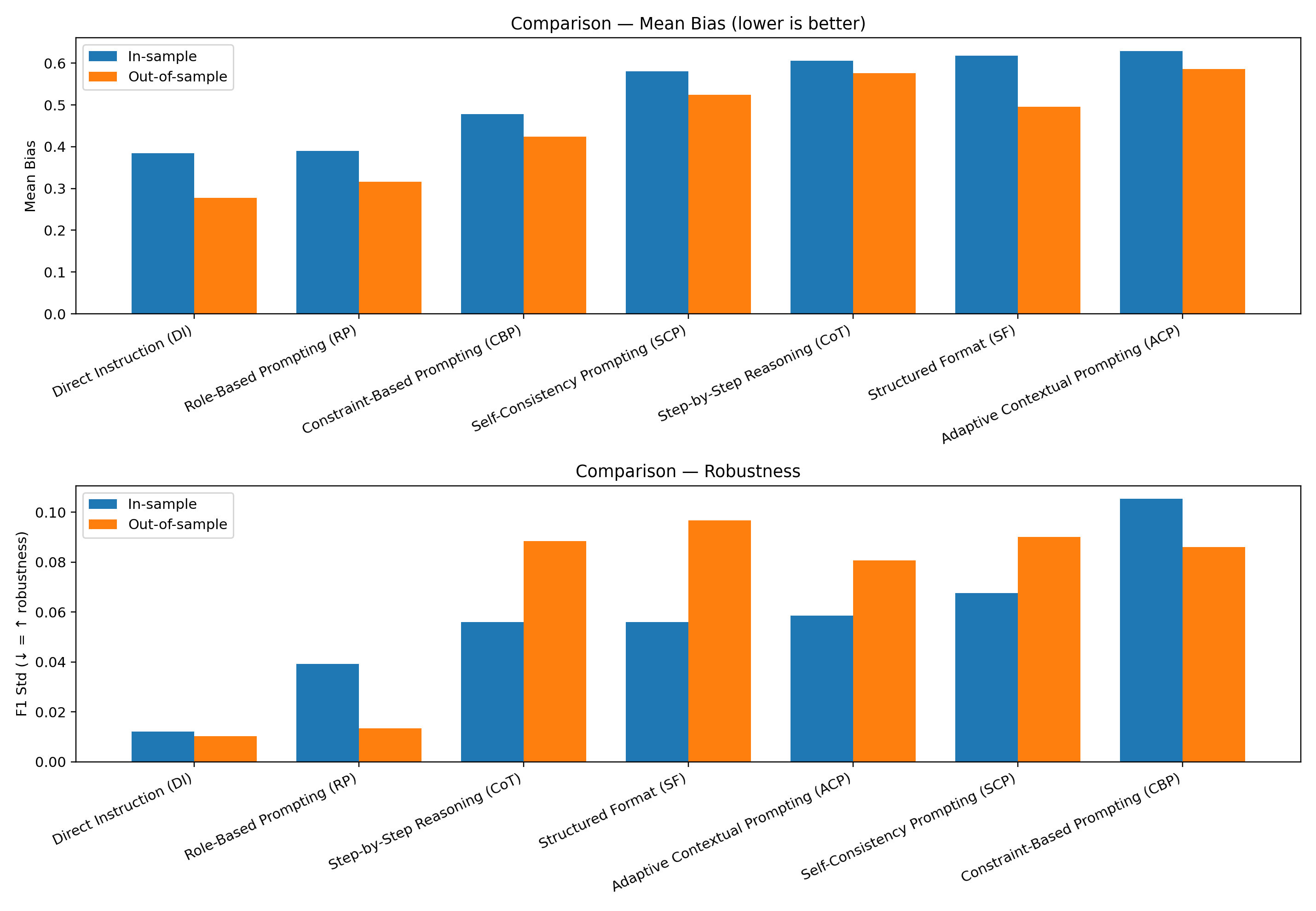}
\caption{IS vs.\ OOS joint comparison of bias and robustness.}
\label{fig:combined_compare_bias_and_robustness}
\vspace{-0.3cm}
\end{figure}

The recommended DI-2 prompt was validated on a 4× larger OOS set. 
Performance remained stable (macro-F1 $\approx$ 0.57, accuracy $\approx$ 0.57). 
Family-wise behavior mirrored IS: DI and RP retained low variance and reduced bias.
Figure~\ref{fig:combined_compare_bias_and_robustness} summarizes IS→OOS shifts, with $\Delta$F1 < 0.03 and minimal $\Delta$Bias/$\Delta\sigma_{F1}$, confirming generalization and behavioral stability.

\subsection{Discussion}
\textbf{(D1)}~\emph{Simplicity over reasoning chains.} DI/RP outperform complex schemes (CoT/ACP), whose multi-step reasoning amplifies context sensitivity.  
\textbf{(D2)}~\emph{Recall as safeguard.} F1-driven selection implicitly privileges high recall, aligning with clinical-screening ethics (avoid false negatives).  
\textbf{(D3)}~\emph{Bias-robustness balance.} The metric triplet $\{F1, |P-R|, \sigma_{F1}\}$ exposes actionable trade-offs between fairness and stability.

\subsection{Lessons for Agentic Engineering}
\textbf{(L1)}~Modular orchestration acts as behavioral assurance—same configuration $\Rightarrow$ same outcome..  
\textbf{(L2)}~Prioritize behavioral stability before architectural complexity.  
\textbf{(L3)}~Integrate \{F1, Bias, Robustness\} as acceptance criteria to operationalize assurance loops.  
\textbf{(L4)}~Framework generalizes to transcript-based health tasks, enabling governed LLM evaluation at scale.

\section{Conclusion and Outlook}
CPEMH operationalizes agentic behavioral assurance for foundation-model systems applied to transcript-based mental-health screening.  
Through autonomous design–evaluation–selection loops, it turns prompt engineering into a reproducible, auditable process that balances accuracy, robustness, and stability.  
Validated on depression-screening transcripts, it provides a transferable blueprint for extending behavioral assurance to other health-related and conversational AI domains.  

\textbf{Future work.} 
Beyond mental health, the framework can be generalized to any \textit{transcript-based foundation-model system}—for example, medical triage, telehealth dialogue analysis, therapy-support interactions, or even educational and customer-service conversations.  
These domains share similar requirements for explainability, behavioral stability, and assurance under contextual variability.  
Extending CPEMH to such scenarios, with retrieval-augmented prompting, adaptive policy agents, and longitudinal drift tracking, will strengthen its role as a modular layer for continuous behavioral governance in agentic AI systems.

\bibliographystyle{ACM-Reference-Format}
\bibliography{references}

@article{powers2020evaluation,
  author    = {David M. W. Powers},
  title     = {Evaluation: From Precision, Recall and F-Measure to ROC, Informedness, Markedness and Correlation},
  journal   = {Journal of Machine Learning Technologies},
  volume    = {2},
  number    = {1},
  pages     = {37--63},
  year      = {2020},
  url       = {https://arxiv.org/abs/2010.16061}
}

@inproceedings{seneviratne2022bias,
  author    = {Suranga Seneviratne and Yong Zhang and Nitin Vaidya and Munindar P. Singh},
  title     = {Bias and Fairness in Artificial Intelligence Systems for Mental Health},
  booktitle = {Proceedings of the 2022 ACM Conference on Fairness, Accountability, and Transparency (FAccT)},
  year      = {2022},
  pages     = {1--12},
  doi       = {10.1145/3531146.3533191}
}

@inproceedings{gratch2014daic,
  author    = {Jonathan Gratch and Ron Artstein and Gale Lucas and Giota Stratou and Stefan Scherer and Angela Nazarian and Rachel Wood and Jill Boberg and David DeVault and Stacy Marsella and David Traum and Skip Rizzo and Albert Morency},
  title     = {The Distress Analysis Interview Corpus of Human and Computer Interviews},
  booktitle = {Proceedings of the Ninth International Conference on Language Resources and Evaluation (LREC)},
  year      = {2014},
  pages     = {3123--3128},
  url       = {http://www.lrec-conf.org/proceedings/lrec2014/pdf/508_Paper.pdf}
}

@article{wang2025pst,
  title={Prompt-Strategy Trees for Multi-Agent Collaboration},
  author={Wang, Haoran and others},
  journal={Proceedings of AAAI 2025},
  pages     = {15621--15629},
  year={2025}
}

@article{jin2024agentmd,
  title={AgentMD: Clinical decision-making agents for medical reasoning},
  author={Jin, Bo and others},
  journal={npj Digital Medicine},
  volume  = {7},
  number  = {1},
  year={2024}
}

@article{yang2025medaide,
  title={MedAide: Agentic reasoning for clinical triage and recommendation},
  author={Yang, Xinyu and others},
  journal={IEEE Transactions on Biomedical Engineering},
  pages   = {1142--1155},
  year={2025}
}

@article{hughes2025agents,
  title={{AI} agents and agentic systems: A multi-expert analysis},
  author={Hughes, L. and Dwivedi, Y.K. and Malik, T. and Shawosh, M. and Albashrawi, M.A. et al.},
  journal={Journal of Computer Information Systems},
  volume={65},
  number={4},
  pages={489--517},
  year={2025}
}

@inproceedings{zhao2024expel,
  title={Expel: {LLM} agents are experiential learners},
  author={Zhao, A. and Huang, D. and Xu, Q. and Lin, M. and Liu, Y.-J. and Huang, G.},
  booktitle={Proceedings of the AAAI Conference on Artificial Intelligence},
  volume={38},
  number={17},
  pages={19632--19642},
  year={2024},
  organization={AAAI}
}

@article{dong2024selfcollab,
  title={Self-collaboration code generation via {ChatGPT}},
  author={Dong, Y. and Jiang, X. and Jin, Z. and Li, G.},
  journal={ACM Transactions on Software Engineering and Methodology},
  volume={33},
  number={7},
  pages={1--26},
  year={2024}
}

@article{toprani2025llmworkflow,
  title={LLM agentic workflow for automated vulnerability detection and remediation in infrastructure-as-code},
  author={Toprani, D. and Madisetti, V.K.},
  journal={IEEE Access},
  volume={13},
  pages={69175--69181},
  year={2025}
}

@inproceedings{jin2024teach,
  title={Teach AI how to code: Using large language models as teachable agents for programming education},
  author={Jin, H. and Lee, S. and Shin, H. and Kim, J.},
  booktitle={Conference on Human Factors in Computing Systems},
  pages={1--28},
  year={2024},
  organization={ACM}
}

@article{qiu2024llmhealth,
  title={{LLM}-based agentic systems in {m}edicine and {h}ealthcare},
  author={Qiu, J. and Lam, K. and Li, G. and Acharya, A. and Wong, T.Y. and Darzi, A. and Yuan, W. and Topol, E.J.},
  journal={Nature Machine Intelligence},
  volume={6},
  number={12},
  pages={1418--1420},
  year={2024}
}

@inproceedings{sasaki2024patterns,
  title={Systematic literature review of prompt engineering patterns in software engineering},
  author={Sasaki, Y. and Washizaki, H. and Li, J. and Sander, D. and Yoshioka, N. and Fukazawa, Y.},
  booktitle={Proceedings of the IEEE 48th Annual Computers, Software, and Applications Conference (COMPSAC)},
  pages     = {670--675},
  year={2024}
}

@inproceedings{brown2020language,
  title={Language models are few-shot learners},
  author={Brown, T.B. and Mann, B. and Ryder, N. and Subbiah, M. and Kaplan, J. and Dhariwal, P. and others},
  booktitle={Advances in Neural Information Processing Systems},
  volume={33},
  pages={1877--1901},
  year={2020}
}

@inproceedings{wei2022chain,
  title={Chain-of-thought prompting elicits reasoning in large language models},
  author={Wei, J. and Wang, X. and Schuurmans, D. and Bosma, M. and Ichter, B. and others},
  booktitle={Advances in Neural Information Processing Systems},
  volume={35},
  pages     = {24824--24837},
  year={2022}
}

@book{white2024patterns,
  title={Chat{GPT} Prompt Patterns for Improving Code Quality, Refactoring, Requirements Elicitation, and Software Design},
  author={White, J. and Hays, S. and Fu, Q. and Spencer-Smith, J. and Schmidt, D.C.},
  publisher={Springer Nature},
  pages = {49--61},
  year={2024}
}

@inproceedings{zhou2023humanprompt,
  title={Large language models are human-level prompt engineers},
  author={Zhou, Y. and Muresanu, A.I. and Han, Z. and Paster, K. and Pitis, S. and Chan, H. and Ba, J.},
  booktitle={International Conference on Learning Representations (ICLR)},
  pages     = {1--24},
  year={2023}
}

@article{son2025optimizing,
  title={Optimizing large language models: A deep dive into effective prompt engineering techniques},
  author={Son, M. and Won, Y.-J. and Lee, S.},
  journal={Applied Sciences},
  volume={15},
  number={3},
  pages   = {1042},
  year={2025}
}

@article{choi2025efficient,
  title={Efficient prompt optimization for relevance evaluation via LLM-based confusion matrix feedback},
  author={Choi, J.},
  journal={Applied Sciences},
  volume={15},
  number={9},
  pages   = {5198},
  year={2025}
}

@inproceedings{zhao2021calibrate,
  title     = {Calibrate Before Use: Improving Few-Shot Performance of Language Models},
  author    = {Tony Z. Zhao and others},
  booktitle = {Proceedings of the 38th International Conference on Machine Learning},
  volume    = {139},
  pages     = {12697--12706},
  year      = {2021},
  series    = {Proceedings of Machine Learning Research},
  publisher = {PMLR},
  url       = {http://proceedings.mlr.press}
}

@article{perez2021truefewshot,
  title={True few-shot learning with language models},
  author={Perez, E. and Kiela, D. and Cho, K.},
  booktitle = {Advances in Neural Information Processing Systems},
  volume    = {34},
  pages     = {11054--11070},
  year={2021}
}


\end{document}